\documentclass[conference]{IEEEtran}
\IEEEoverridecommandlockouts
\usepackage{cite}
\usepackage{amsmath,amssymb,amsfonts}
\usepackage{algorithmic}
\usepackage{graphicx}
\usepackage{textcomp}
\usepackage{xcolor}
\usepackage{url}
\def\BibTeX{{\rm B\kern-.05em{\sc i\kern-.025em b}\kern-.08em
    T\kern-.1667em\lower.7ex\hbox{E}\kern-.125emX}}
\begin{document}

\title{Understanding Driving Risks using Large Language Models: Toward Elderly Driver Assessment\\
\author{
\IEEEauthorblockN{
Yuki Yoshihara\IEEEauthorrefmark{1},
Linjing Jiang\IEEEauthorrefmark{1},
Nihan Karatas\IEEEauthorrefmark{1},
Hitoshi Kanamori\IEEEauthorrefmark{1},
Asuka Harada\IEEEauthorrefmark{1},
Takahiro Tanaka\IEEEauthorrefmark{1}
\IEEEauthorblockA{\IEEEauthorrefmark{1}Institutes of Innovation for Future Society, Nagoya University, Japan}
}
}
}
\maketitle

\begin{abstract}
This study investigates the potential of a multimodal large language model (LLM), specifically ChatGPT-4o, to perform human-like interpretations of traffic scenes using static dashcam images. Herein, we focus on three judgment tasks relevant to elderly driver assessments: evaluating traffic density, assessing intersection visibility, and recognizing stop signs recognition. These tasks require contextual reasoning rather than simple object detection. Using zero-shot, few-shot, and multi-shot prompting strategies, we evaluated the performance of the model with human annotations serving as the reference standard. Evaluation metrics included precision, recall, and F1-score. Results indicate that prompt design considerably affects performance, with recall for intersection visibility increasing from 21.7\% (zero-shot) to 57.0\% (multi-shot). For traffic density, agreement increased from 53.5\% to 67.6\%. In stop-sign detection, the model demonstrated high precision (up to 86.3\%) but a lower recall (approximately 76.7\%), indicating a conservative response tendency. Output stability analysis revealed that humans and the model faced difficulties interpreting structurally ambiguous scenes. However, the model’s explanatory texts corresponded with its predictions, enhancing interpretability. These findings suggest that, with well-designed prompts, LLMs hold promise as supportive tools for scene-level driving risk assessments. Future studies should explore scalability using larger datasets, diverse annotators, and next-generation model architectures for elderly driver assessments.

\end{abstract}


\bibliographystyle{IEEEtran}

\section{INTRODUCTION}
Intersections present daily risks for elderly drivers. Each year, approximately 175,000 crashes occur at intersections, with drivers aged 65 and older involved in 60\% of these incidents \cite{npa2024traffic}. Combined with a fatality rate of 0.7\%, which is comparable to the overall average for traffic accidents, this high incident rate represents a considerable safety concern.

Providing drivers with objective diagnostic reports, which summarizes the types of intersections encountered and their corresponding responses, can promote safer driving \cite{romoser2013long, bell2017evaluation, castellucci2020interventions}. Although such feedback has been proven to improve driver awareness \cite{picco2023use}, its generation remains largely dependent on human judgment, with limited progress made toward automation \cite{tanaka2018study,driessen2024ai}.

\begin{figure}[t]
    \centering
    \includegraphics[scale=0.3]{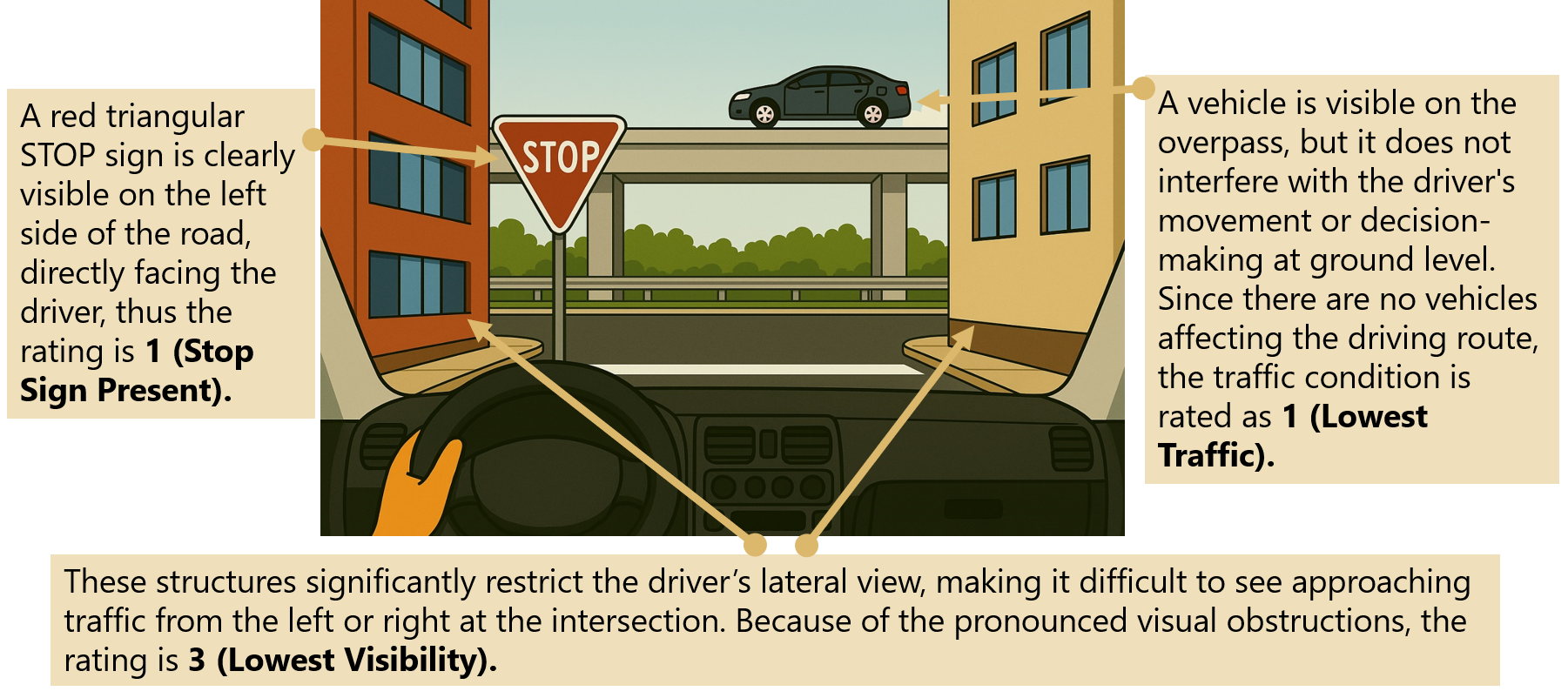}
    \caption{Conceptual illustration of the study. This figure simulates the type of relational reasoning that large language models (LLMs) perform--—assigning risk labels not only based on the visual presence of objects, but on their contextual relevance to the driving behavior of the ego-vehicle. For example, a stop sign is only relevant if it faces the driver. Vehicles separated by barriers pose minimal risk, whereas buildings that block the side view of the driver introduce latent risks by reducing intersection visibility.}
    \label{fig:concept}
\end{figure}

A key challenge in automating traffic situation assessment is relational judgment, determining the relevance of each element with respect to the ego-vehicle. Unlike in simple object detection, this requires understanding the context, spatial relationships, and intent. When assessing traffic volume for example, detecting vehicles in an image does not accurately reflect risk. The risk must be assessed based on the position and intent of the ego-vehicle. Vehicles visible in the frame but separated by physical barriers from merging into the ego lane should be considered low risk. This aligns with the kind of judgment human drivers routinely apply. Similarly, assessing intersectional visibility presents challenges; risk must not be assessed based solely on visible hazards but also on potential threats, such as pedestrians or vehicles that could emerge unexpectedly. Even for traffic signs, the key is not their presence in the image, but whether they are directed toward the ego-vehicle. For example, a stop sign facing away from or laterally to the driver is irrelevant to the driving scenario (Fig. \ref{fig:concept}).

Recent advances in large language models (LLMs) have enabled the computational integration of relational information \cite{zhang2024integrating,charoenpitaks2024exploring}. Tools such as ChatGPT-4o offer user-friendly APIs that go beyond language completion, supporting text generation from structured data, images, and videos. 
However, although prior studies have explored the generative capabilities of LLMs for driving-related tasks, such as behavior simulation \cite{jin2024surrealdriver}, natural language explanation \cite{xu2024drivegpt4}, and conversational support \cite{huang2024chatbot}, there remains a critical gap in structured, quantitative evaluation of driving scenes. Specifically, existing work does not address how LLMs can assess scene-level risk factors (e.g., intersection visibility, traffic volume, and stop-sign relevance), which are essential for objective and interpretable driver diagnostics, particularly for elderly drivers \cite{lerner1995older,keay2009urban,swain2021left}. Hence, this study investigates the feasibility of using these models to support automated assessment of traffic scenarios based on their relevance to the driver.

In this study, we extracted still images from dashcam videos of elderly drivers and used ChatGPT-4o, a multimodal generative AI, to describe these scenes. Although video-based interpretation remains a long-term goal, we focus on still images owing to the current limitations of multimodal models, which process static images more reliably and efficiently than video streams. Given our interest in assessing elderly driving behavior, we prompted the model to evaluate traffic context features that rely on human on judgment, such as intersection visibility, traffic volume, and stop sign presence, and assess the plausibility of its outputs.

\subsection{Research Question}

This study addresses a central question: To what extent can multimodal generative AIs emulate the integrated judgment of human drivers? We investigated whether current state-of-the-art AI models can accurately label traffic conditions that are crucial for elderly driver diagnostics, such as traffic density, intersection visibility, and stop-sign presence, based on their relevance to ego vehicles. These questions are critical for advancing the automation of driver diagnostics using driver-recorder data.

\subsection{Contributions of this Study}
\begin{enumerate}
\item Prompt Design for Traffic Scene Interpretation: We investigate the effect of different prompting strategies, zero-shot \cite{kojima2022large}, few-shot \cite{brown2020language}, and multi-shot, on the ability of ChatGPT-4o to assess structured risk elements in traffic scenes. By comparing these methods, we determine the degree of contextual guidance required for accurate and consistent assessments. The results offer insights into effective prompt construction, with implications for AI-assisted driving assessment and scene interpretation.

\item Human--AI Judgment Comparison: We compare AI-generated labels with human ratings to evaluate how closely the model approximates human risk perception in traffic scenes. This benchmarking highlights areas of alignment and divergence between the model and human reasoning, thereby offering insights into the current reliability of the model and the continued importance of human oversight in safety-critical tasks such as elderly driver evaluation and traffic risk analysis.
\end{enumerate}

\section{RELATED WORKS}
The application of large language models (LLMs) in driving contexts has grown rapidly in recent years, with numerous studies investigating their potential from different perspectives.

Xu et al. (DriveGPT4) proposed a method for predicting vehicle behavior (steering and speed) during autonomous driving, while generating natural language explanations of these actions to improve the interpretability of otherwise opaque control systems \cite{xu2024drivegpt4}.

Park et al. introduced a benchmark dataset for natural language question answering (QA) on driving scenes, which incorporates multi-view and bird’s-eye view (BEV) visual features to enable scene understanding from different perspectives \cite{park2025nuplanqa}.

Huang et al. demonstrated the potential of LLM-based chatbots in supporting driving performance \cite{huang2024chatbot}. Through real-time interactions with drivers, the chatbot reduced fatigue and enhanced attentiveness, thereby highlighting the effectiveness of conversational AI for driving support.

Jin et al. (SurrealDriver) proposed a driver agent that simulated human driving behavior \cite{jin2024surrealdriver}. They leveraged natural language prompts from human thought processes while driving to guide the behavior of an LLM-based generative agent.

These studies focused primarily on the generative capabilities of LLMs—--natural language generation, behavior generation, and dialogue generation—--aimed at producing human-readable descriptions or actions based on driving scenes.

Charopenpitaks et al. proposed a framework for predicting and explaining potential hazards using a single dashcam image, emphasizing abductive reasoning (inferring what might happen next in a driving scene) \cite{charoenpitaks2024exploring}. Our study shares a similar foundational concept, as both approaches seek to extract driving-relevant information from single static images using LLMs. The aforementioned is rooted in the current limitations of LLMs, which lack robust support for continuous video inputs.

However, both their study and the aforementioned studies focus on generating free-form descriptions, predicted actions, or interactive dialogue, without addressing the structured, quantitative evaluation of driving scenes.
In particular, questions such as the following:
“How limited is the visibility at this intersection?”, “What is the traffic volume?”, and “Is a stop required here?” remain unanswered, yet are critical for a structured diagnostic evaluation of the driving behavior, notably in elderly drivers \cite{lerner1995older,keay2009urban,swain2021left}.

Human drivers tend to accept evaluations by machines or experts, particularly when the results are presented in a neutral and structured format \cite{roetting2003technology, picco2023use}. Hence, ensuring the accuracy of machine-based assessment is not only crucial scientifically but also an ethical requirement for the responsible use of AI in safety-critical applications. In the following sections, the design and methodology used to address this problem is discussed.

\section{METHOD}
\subsection{Overview}
\begin{figure}[t]
    \centering
    \includegraphics[scale=0.35]{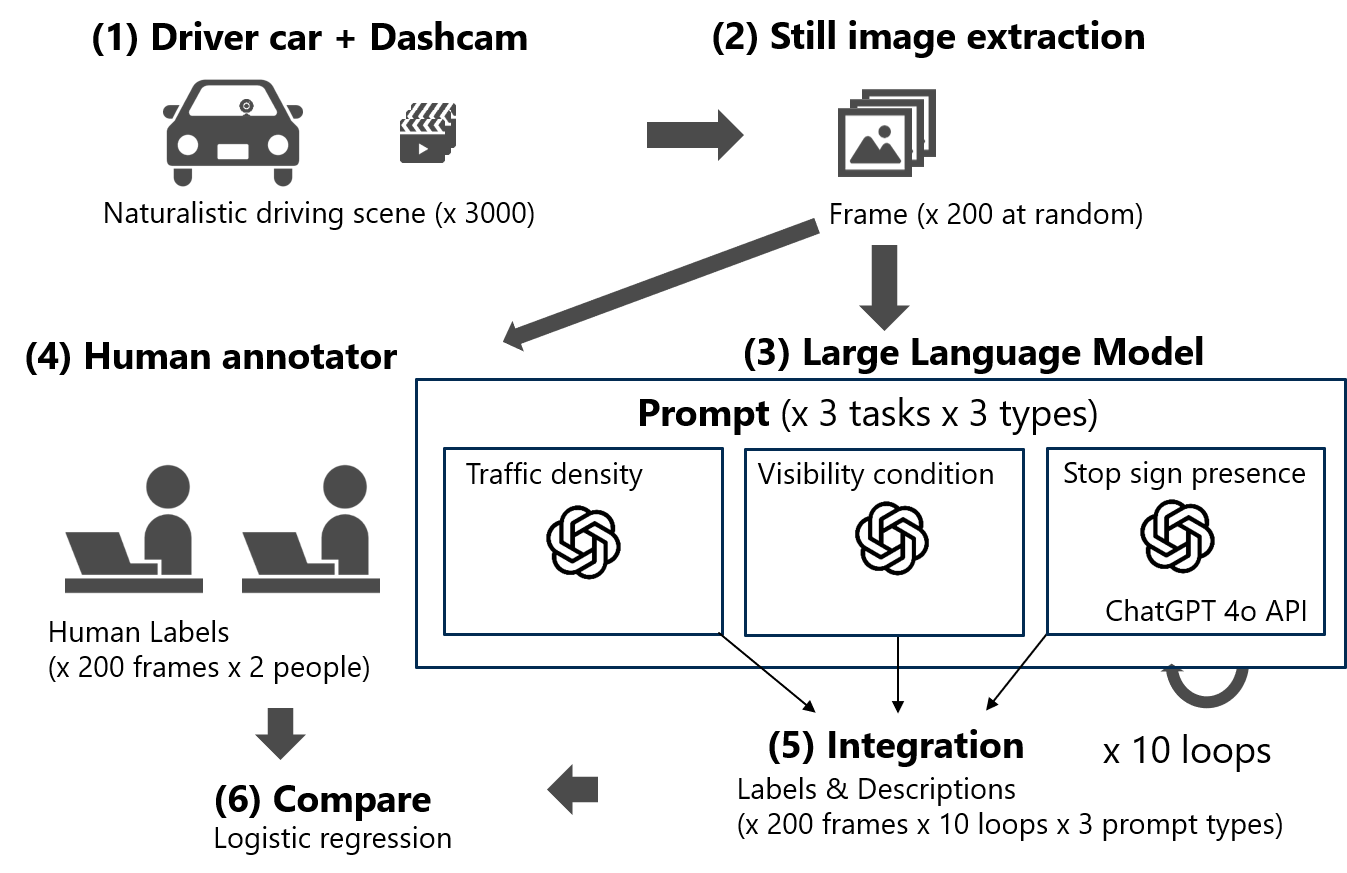}
    \caption{Evaluation pipeline.}
    \label{fig:pipeline}
\end{figure}

The evaluation pipeline in this study is illustrated in Fig. \ref{fig:pipeline}. First, dashcam video frames were extracted and used to develop a structured image dataset (1--2). Each image was then evaluated using ChatGPT-4o (3) and two human raters (4) across three classification tasks: traffic density, intersection visibility, and stop-sign presence. To ensure clarity and modularity, we assigned each task a separate prompt and the model was tested using three prompting strategies: zero-shot, few-shot Chain-of-Thought (CoT), and multi-shot CoT. The model outputs were programmatically integrated (5) and compared to human annotations (6).

\subsection{Dashcam Dataset}
We used image frames extracted from a dashcam video corpus collected in a previously approved naturalistic driving study. Although the dataset is not publicly available, it was carefully curated and anonymized to ensure privacy and reflect the real-world driving risks faced by elderly drivers. The full dataset comprised approximately 3,000 scene-level video segments (640×360 pixels) primarily captured on rural roads and frequently involving intersections (scenarios known to present elevated risks to older drivers).

From the aforementioned, we randomly selected approximately 200 representative scenes, using one frame per scene to avoid redundancy. Although numerous images were initially considered, we limited the dataset to 200 owing to two constraints: (1) API limits on ChatGPT-4o restrict the number of feasible image-based queries and (2) our analysis required 10 repeated generations per image to evaluate the response stability. These factors made the 200 input images practical.

\subsection{Prompt Design for ChatGPT-4o}\label{subsec:prompt}
\begin{figure}[t]
    \centering
    \includegraphics[scale=0.4]{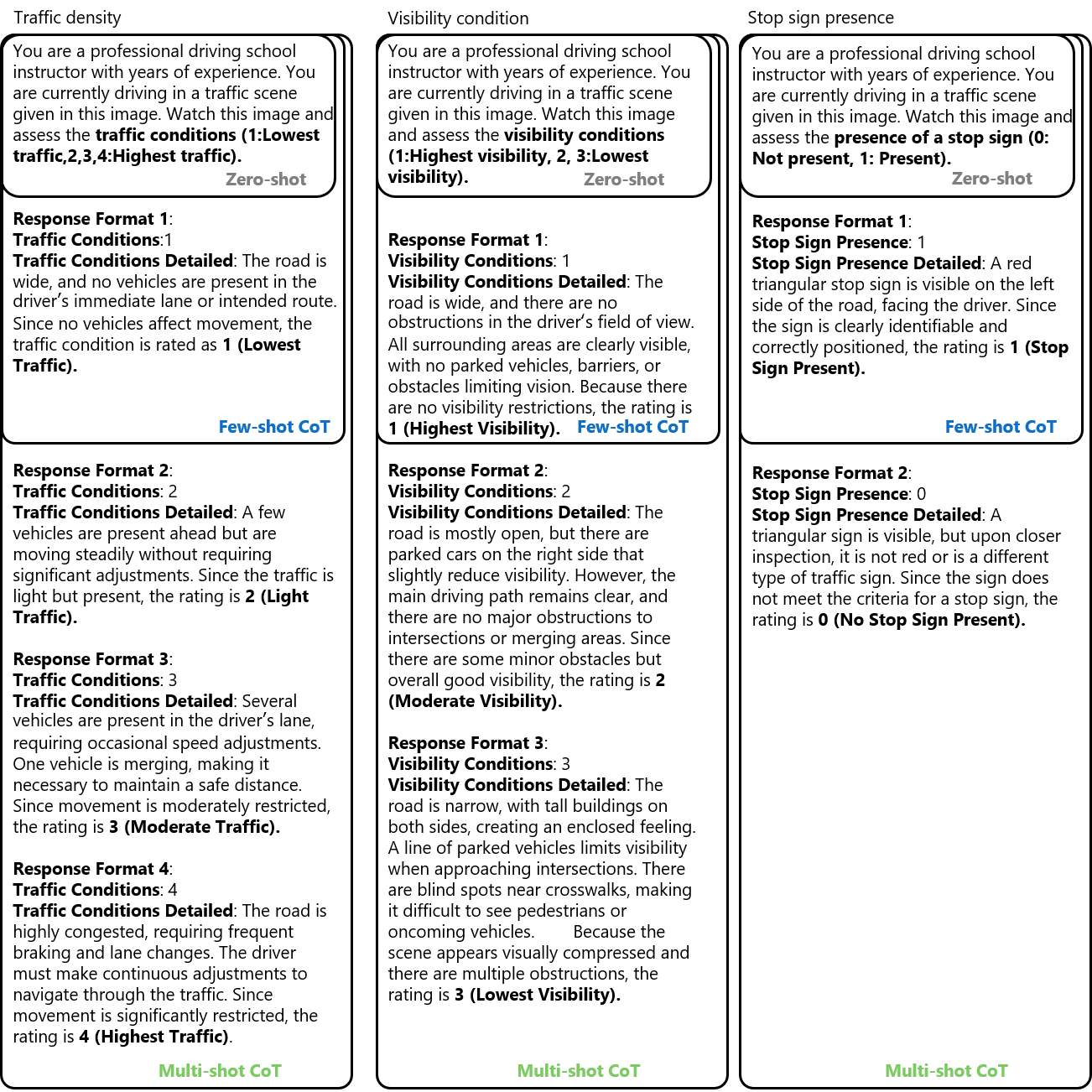}
    \caption{Prompts used for each classification tasks: Zero-shot (top), few-shot CoT (middle), and multi-shot CoT (bottom) share a similar response format but differ in the number of examples provided.}
    \label{fig:prompt}
\end{figure}

Figure \ref{fig:prompt} shows the nine prompt templates, spanning three tasks (traffic conditions, intersection visibility, and stop sign presence) and three strategies: (1) zero-, (2) few-, and (3) multi-shot. Each task was prompted separately for clarity. The three strategies differed only in the number of label–explanation examples provided, while maintaining a consistent reasoning-based format. In all the cases, the model categorized traffic scenes along the same three dimensions.

In zero-shot prompting, the prompt includes the role, task description, and expected response format without examples provided. In few-shot CoT prompting, one or two exemplar categories were included, each paired with a concise natural-language explanation outlining the conditions for labeling. In multishot CoT prompting, the model listed all possible categories for a given task (traffic levels 1–-4), each accompanied by a detailed explanation to serve as a comprehensive reference.

The aforementioned three configurations were selected to investigate the effect of varying-levels of category-specific guidance on the accuracy of the model in structured classification tasks. The model must assign semantically distinct labels based on contextual understanding; hence, we hypothesized that providing multiple exemplars would improve consistency and interpretability. Simultaneously, we aimed to assess whether the pretraining of the model on vision–-language data would support reasonable performance even without explicit examples.
Hence, zero-shot prompting was used as a baseline for comparison.

All prompt types followed a consistent layout, with the model instructed to respond using a predefined output format (visibility: [highest/moderate/lowest]). The CoT explanations embedded in the exemplars aim to simulate analogical reasoning, thereby guiding the model to align new scenes with semantically appropriate reference cases.

\subsection{Super-parameters for ChatGPT-4o}
We used OpenAI GPT-4o (gpt-4o-2024-11-20) for all scene classification tasks using the OpenAI Python API between December 2024 and February 2025. The model was queried using chat.completions.create with the following settings: temperature = 0.2, top p = 0.5, maximum tokens = 2000, frequency penalty = 0.3, and presence penalty = 0.2. These parameters were selected to ensure accuracy and consistency, while minimizing the output variability for structured labeling. The prompts followed a standardized message format (messages=$\textrm{[PROMPT\_MESSAGES]}$), while the responses were parsed automatically using a Python script. No system messages or special role conditioning were used beyond the prompt content.

\subsection{Human Annotation Protocol}
To evaluate the generative AI output, we assigned two human raters (experienced drivers with relatively cautious driving styles) the task of labeling a set of dashcam images. Aside from being provided with a common evaluation manual, they worked independently without interaction or collaboration to minimize bias.
They evaluated each image using the following core dimensions:
\begin{itemize}
\item Traffic density: Rated as Lowest, Light, Moderate, or Highest.
\item Intersection visibility: Rated as Highest, Moderate, or Lowest.
\item Stop signage presence: Rated Present or Not present, a binary label based on whether a red triangular stop sign is clearly visible and oriented toward the ego-vehicle.
\end{itemize}
They were instructed to adopt the perspective of the driver and assess the scene as if they were actively driving. This perspective-based instruction corresponded with the format used in \ref{subsec:prompt} when generative AI was prompted to ensure consistency between the human and model evaluations.

\subsection{Performance Metrics: Human vs. Model}
To assess the model performance, we compared the labels generated by ChatGPT-4o with human annotations using a categorical accuracy-based metric. For each task, the prediction of the model was correct (1.0) if it corresponded with the human-annotated label, and incorrect (0.0) otherwise. 

The evaluation was conducted using 200 unique images, each labeled independently by two human raters, thereby resulting in 400 annotations. The model was applied to the same 200 images using 10 repeated runs per prompt configuration, thereby generating 2,000 model outputs. For direct comparison, each of the 400 human annotations was replicated ten times, thereby resulting in 4,000 human label entries, which were then paired with the 4,000 corresponding model outputs. In this setup, we treated each label of the rater independently, without aggregating or averaging across raters.

The proportion of correct matches was computed as the primary accuracy metric and was calculated separately for each prompting (zero-shot, few-shot CoT, multi-shot CoT) across the three classification tasks: traffic conditions, intersection visibility, and stop-sign presence. 

To evaluate the impact of the prompting strategy on the labeling accuracy, we conducted logistic regression analysis using a generalized linear model with a binomial distribution and logit link function. The binary response variable (y) indicated whether the prediction of the model corresponded with the human label (1=correct and 0=incorrect), while the predictor variable (condition) represented the prompting strategy used (zero-shot, few-shot CoT, multi-shot CoT).

The model was implemented using MATLAB’s fitglme function, while the statistical significance of the prompt condition was assessed via an overall F-test using coefTest(glme). To interpret the results in terms of success probabilities, we converted the model estimates from a logit to response scale using an inverse logit function. We calculated the predicted success probability for each prompt as $p = {e^\eta}/{(1 + e^\eta)}$, where $\eta$ denotes a linear predictor for each condition. Further, a corresponding 95\% confidence interval was obtained by applying a similar transformation to the lower and upper bounds of each coefficient. These probabilities were used to summarize and visualize the effect of the prompting strategy on the model performance. A significance threshold of p $<$ 0.05 was used throughout, while the statistical analyses were performed using MATLAB R2022b.

\subsection{Ethical note}
We conducted a secondary analysis of anonymized dashcam videos using a previously IRB-approved study. No identifiable information was used, and no new human-subject interventions were involved.
Human annotation was conducted by external contractors who were compensated for their work. The study was approved by the Institutional Review Board of Nagoya University, Japan (Approval No. 2024-2).

\section{RESULT}

\subsection{Representative Labels}
\begin{figure}[t]
    \centering
    \includegraphics[scale=0.33]{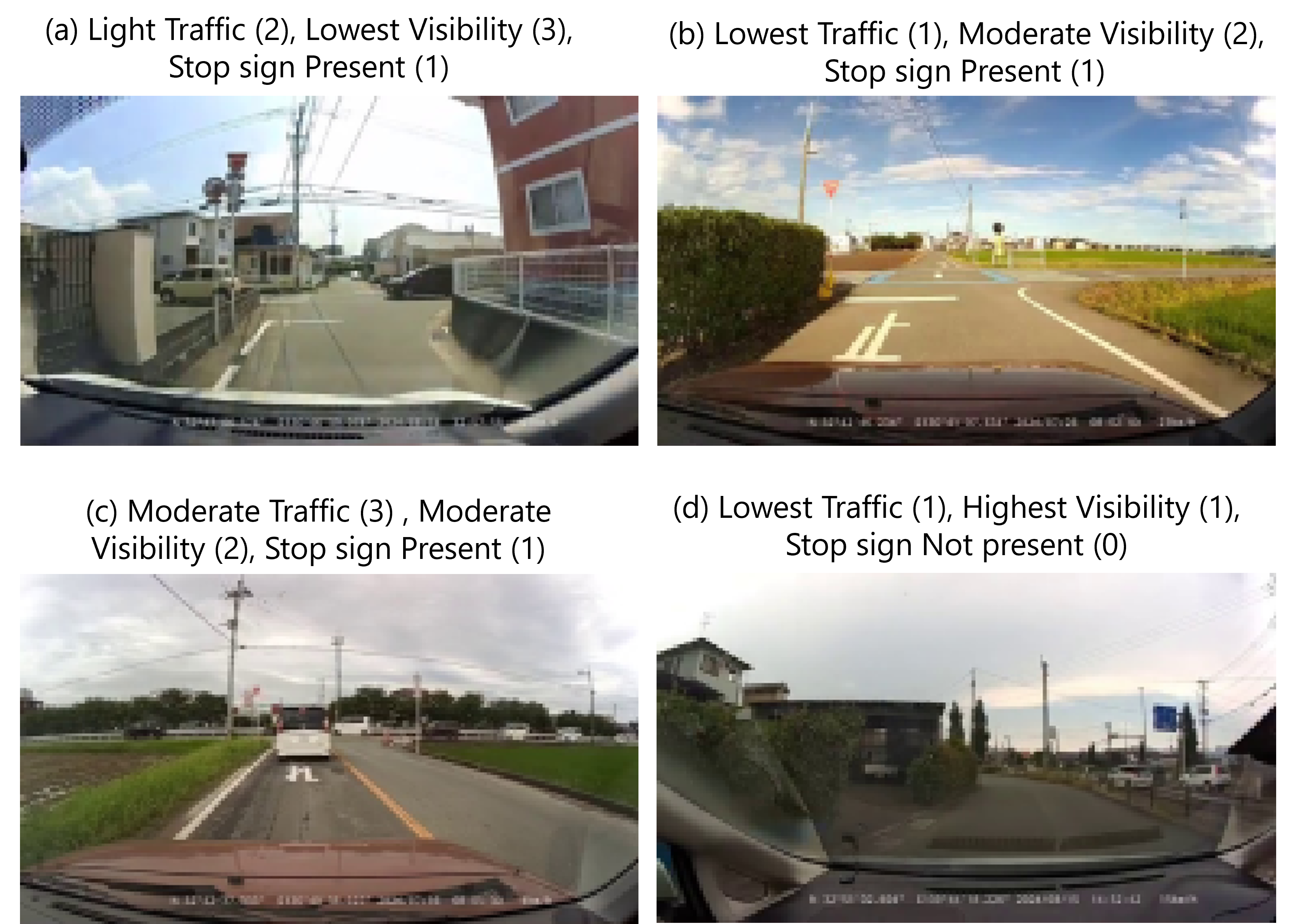}
    \caption{Representative dashcam images and example outputs using the LLM under different prompting conditions. Certain regions of the images have been blurred to ensure privacy.}
    \label{fig:samples}
\end{figure}

Figure \ref{fig:samples} shows the representative label predictions by ChatGPT-4o using different prompting strategies, each based on forward-facing dashcam images.

Figure \ref{fig:samples}(a) shows a vehicle approaching the main road from a narrow street with buildings obstructing lateral visibility. A stop sign is clearly visible, and the two vehicles travel across the street. ChatGPT labeled the scene as light traffic (2), lowest visibility (3), and stop sign present (1), which closely corresponded with the labels assigned by both raters, who labeled stop sign as present (1), visibility as lowest (3), and traffic as light (2) and moderate (3), respectively.

Figure \ref{fig:samples}(b) shows a rural intersection with a partial left-side obstruction from vegetation and no other vehicles. ChatGPT labeled the scene as having the lowest traffic (1), moderate visibility (2), and stop sign present (1), which corresponded with the labels assigned by the raters.

Figure \ref{fig:samples}(c) shows numerous vehicles on a cross street and one directly ahead of the driver. ChatGPT outputs were moderate traffic (three), moderate visibility (two), and stop signs (one). Raters agreed with traffic (3) and stop sign (1). For visibility, one rater agreed with Model (2), while the other rated it as lowest (3).

Figure \ref{fig:samples}(d) shows a side road parallel to the main road separated by a fence. Under the few-shot CoT prompt, ChatGPT labeled traffic as lowest (1) and explained: “The road is clear, with no vehicles obstructing the immediate lane of the driver or intended route...” Visibility was labeled highest (1) and stop sign as not present (0). Raters agreed on traffic (1) and stop signs (0) but rated visibility as moderate (2), thereby indicating that the model overestimated visibility.

For the multishot prompt, ChatGPT labeled a similar scene as light traffic (2) with the following reason: “A few vehicles are visible on the road ahead... they do not require considerable adjustments.” However, this overlooked separation of the road by a fence, thereby suggesting that more examples in a prompt do not always yield better alignment. These examples demonstrate cases where the structured output of ChatGPT and accompanying explanations either correspond with or diverge from human judgments. The following section presents statistical comparisons across the prompting strategies.

\subsection{Impact of Prompt Type on Labeling Accuracy}
\begin{figure}[t]
    \centering
    \includegraphics[scale=0.33]{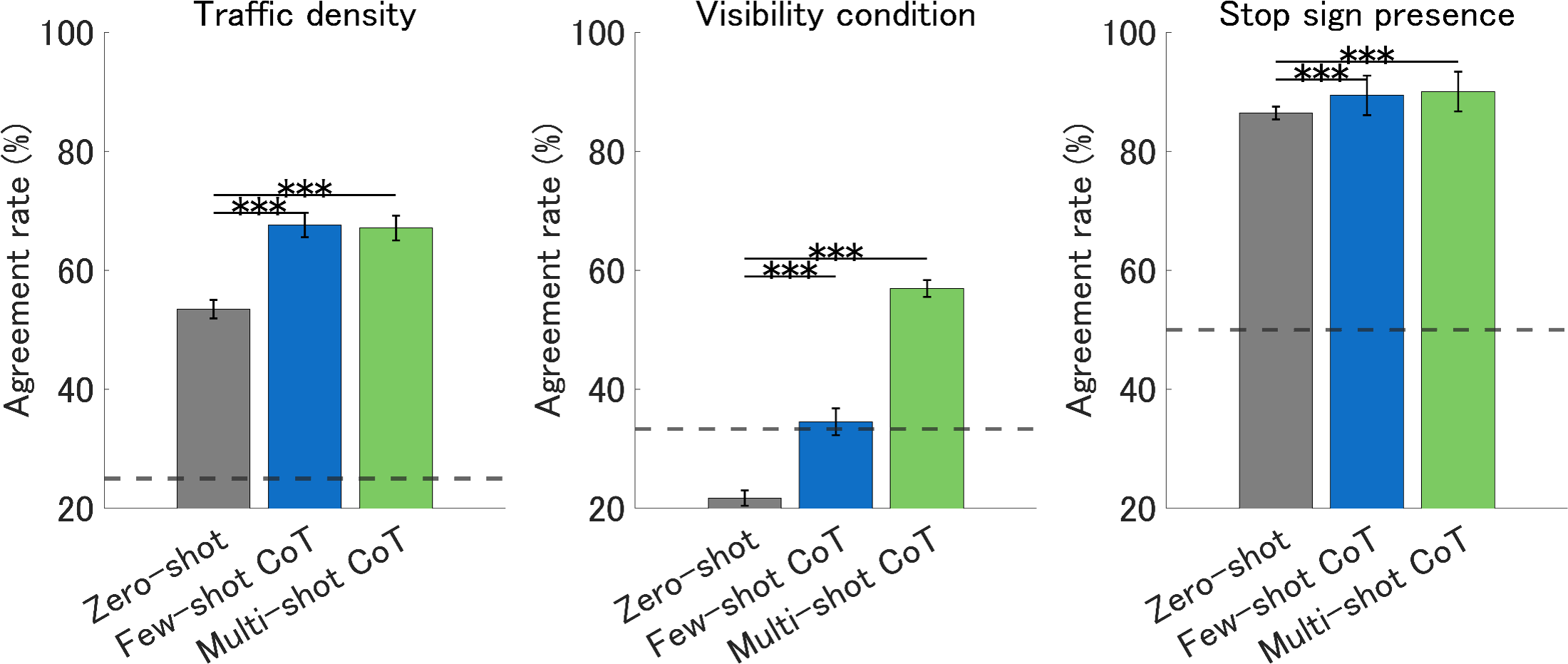}
    \caption{Agreement rates across three prompting strategies (Zero-shot, Few-shot CoT, Multi-shot CoT) for each classification task. Error bars represent 95\% confidence intervals, while dashed lines indicate the chance-level accuracy. The asterisks indicate the statistical significance: p $<$ 0.001.}
    \label{fig:agreement_rate}
\end{figure}

Figure \ref{fig:agreement_rate} shows the agreement rates between the model predictions and human annotations for each prompting strategies evaluated separately for each task. To quantify the effects of prompting, logistic regression and ANOVA were conducted for each classification task.

The estimated agreement probability for traffic density was 53.5\% under the zero-shot prompt. This increased to 67.6\% with the few-shot CoT prompt and to 67.1\% with the multi-shot CoT prompt. ANOVA revealed a considerable impact of the prompting condition, with $F$(2, 11997) = 109.21, $p$ $<$ 0.001. Logistic regression indicated that the few-shot ($\beta$ = 0.60, SE = 0.046, $t$(11997) = 12.89, $p$ $<$ 0.001, 95\% CI [0.51, 0.69]) and multi-shot ($\beta$ = 0.57, SE = 0.046, $t$(11997) = 12.43, $p$ $<$ 0.001, 95\% CI [0.48, 0.67]) considerably improved the agreement over zero-shot. However, multi-shot did not outperform few-shot in this case.

This improvement is more pronounced for intersection visibility. The agreement probability increased from 21.7\% (zero-shot) to 34.6\% (few-shot) and 57.0\% (multishot). ANOVA revealed a robust effect of the prompting condition, $F$(2, 11997) = 512.51, $p$ $<$ 0.001. Regression coefficients confirmed a considerable increase for the few-shot ($\beta$ = 0.64, SE = 0.051, $t$ = 12.67, $p$ $<$ 0.001, 95\% CI [0.54, 0.74]) and multi-shot ($\beta$ = 1.56, SE = 0.050, $t$ = 31.30, $p$ $<$ 0.001, 95\% CI [1.46, 1.66]).

For the stop-sign recognition task, the agreement probability was high under zero-shot (86.5\%) and further improved to 89.4\% (few-shot) and 90.1\% (multi-shot). ANOVA indicated a considerable effect of the prompting condition, $F$(2, 11997) = 14.53, $p$ $<$ 0.001. Regression analysis showed  positive effects for few-shot ($\beta$ = 0.28, SE = 0.069, $t$ = 4.04, $p$ $<$ 0.001, 95\% CI [0.14, 0.41]) and multi-shot ($\beta$ = 0.35, SE = 0.070, $t$ = 4.98, $p$ $<$ 0.001, 95\% CI [0.21, 0.49]).

These results demonstrate that prompt design considerably affected the model performance across all the tasks, with the largest gains observed in the intersection visibility task. However, that the dataset was imbalanced, with a larger number of scenes lacking stop signs compared to those with visible signs. Consequently, the overall accuracy is more reflective of the performance of the majority class. To better understand the behavior of the model across the different categories, the next subsection analyzes the category-specific performance using confusion matrices.

\subsection{Category-Level Performance and Confusion Analysis}
In this section, we focus on the results obtained from multishot (CoT) prompting, which yielded the highest overall accuracy in our previous experiments. To further examine the behavior of the model, we analyzed the distribution of predicted labels across categories and compared them to human annotations.

\begin{figure}[t]
    \centering
    \includegraphics[scale=0.35]{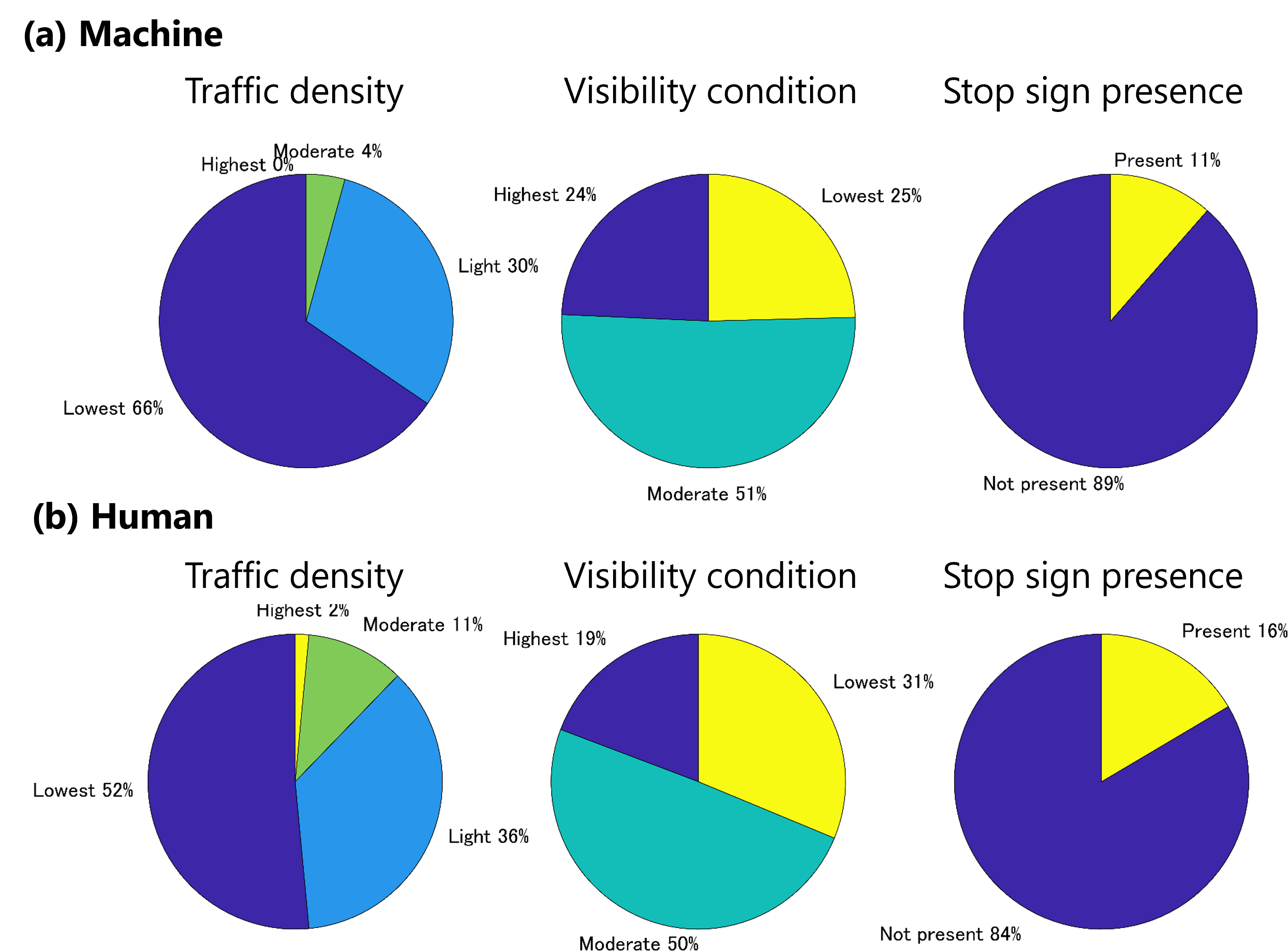}
    \caption{Distribution of categorical labels generated by the model (top row) and human raters (bottom row) for each classification task. }
    \label{fig:charts}
\end{figure}

Figure \ref{fig:charts} shows a comparison of the model-generated label distribution (top row) and human-annotated distribution (bottom row) for three classification tasks: traffic density, visibility condition, and stop-sign presence. Across all tasks, the outputs of the model reflected the global labeling trends of the human raters. For traffic density, the model and humans assigned lower-level labels the most, with relatively fewer cases classified as “Moderate” or “Highest.” For intersection visibility, “Moderate” was the most common label in both distributions, with “Highest” and “Lowest” assigned less often and in comparable proportions. For stop sign presence, both distributions reflected a high prevalence of scenes without stop signage. However, the model slightly overpredicted the “Not present” category compared to the human annotations.

\begin{figure}[t]
    \centering
    \includegraphics[scale=0.35]{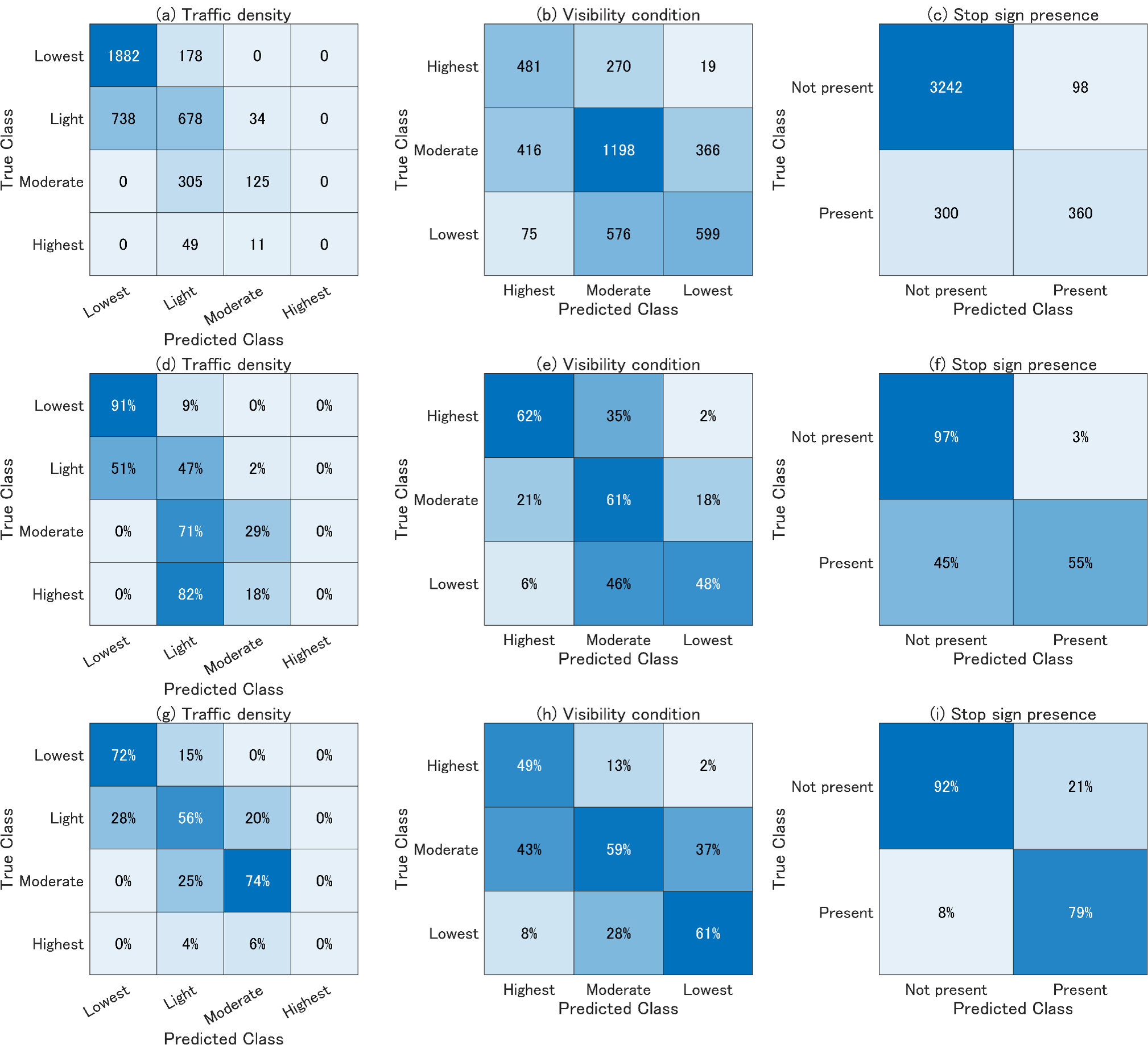}
    \caption{Confusion matrices (top row), recall matrices (middle row), and precision matrices (bottom row) for traffic density (a, d, g), visibility condition (b, e, h), and stop sign presence (c, f, i) under the multi-shot CoT prompting condition. Confusion matrices show the distribution of predicted versus true labels; recall matrices report the proportion of correctly predicted instances within each true class; precision matrices report the proportion of correctly predicted instances within each predicted class.}
    \label{fig:matrix}
\end{figure}

Figure \ref{fig:matrix} shows the confusion matrices (top rows) which correspond with recall matrices (middle rows) and precision matrices (bottom rows) for each classification task under the multi-shot CoT prompting condition. While the confusion matrices provide raw counts of predicted versus true labels, the recall matrices highlight the effectiveness of the model in recovering each true category, notably the proportion of correct predictions out of all the instances for a given true class. Conversely, the precision matrices show the proportion of correct predictions for each predicted class, thereby offering insight into the reliability of the positive predictions of the model.

In this study, we focused primarily on recall because it reflects the ability of the model to detect relevant cases without omissions. This is crucial in diagnostic applications where failure to detect an actual issue, such as a stop-sign or limited visibility, can result in safety risks. However, precision also provides complementary information on the prediction validity, particularly for downstream use cases wherein high-confidence filtering is required.

In the traffic density task, the model achieved the highest recall at 91\%, thereby indicating robust sensitivity to low-traffic scenarios. Conversely, the recall scores for light and moderate traffic were considerably lower at 47\% and 29\%, respectively. The model failed to recall any instances with the highest traffic (0.0\%), thereby highlighting the challenge in detecting denser traffic conditions. However, the precision remained stable across the first three classes (72\%, 56\%, and 74\%, respectively), although that for the highest traffic was 0\%, corresponding with its recall failure.

In the visibility task, the recall scores were 62\%, 61 \%, and 48 \% for the highest, moderate, and lowest scenes, respectively. Further, the precision values in this task were balanced, thereby indicating the relative sensitivity and accuracy of the predictions of the model.

In the stop-sign presence task, the model demonstrated strong recall for non-present scenes (97\%), while the recall for present scenes was lower at 55\%. The class precision for the present category was 79\%, thereby indicating that although the model frequently missed stop signs, its positive predictions were reliable.

\subsection{Consistency across Raters and Model}
To assess the consistency and reliability of the model predictions relative to human judgments, we compared its performance to that of individual raters and their majority agreement. Inter-rater agreements between the human annotators was quantified using Cohen’s kappa, yielding 0.688 for traffic density, 0.325 for visibility conditions, and 0.927 for stop-sign presence. These results indicate a strong agreement for stop-sign annotations, moderate agreement for traffic density, and low agreement for visibility, thereby suggesting that the consistent interpretation of intersection visibility posed challenges even among human raters.

To further investigate the performance consistency, we conducted a systematic comparison across four evaluator types: Rater 1, Rater 2, majority vote (R1 \& R2), and machine. For each classification task, we computed the precision, recall, and F1-score using macro-averaging across classes. These metrics enable direct comparisons under a uniform evaluation criterion. The results are summarized in Table \ref{tbl:performance_metrics}. 

\begin{table}[htb]
\centering
\caption{Pairwise macro-averaged precision, recall, and F1-scores between different annotation sources.}
\label{tbl:performance_metrics}
\begin{tabular}{llllll}
\hline
Task & Ground Truth & Predicted By & Precision & Recall & F1 \\
\hline
Stop Sign & Rater1 & Rater2 & 0.964 & 0.964 & 0.964 \\
Stop Sign & Rater2 & Rater1 & 0.964 & 0.964 & 0.964 \\
Stop Sign & Rater2 & Machine & 0.863 & 0.767 & 0.803 \\
Stop Sign & R1\&R2 & Machine & 0.851 & 0.758 & 0.793 \\
Stop Sign & Rater1 & Machine & 0.838 & 0.749 & 0.783 \\
&&&&&\\
Traffic & Rater2 & Rater1 & 0.656 & 0.815 & 0.663 \\
Traffic & Rater1 & Rater2 & 0.815 & 0.656 & 0.663 \\
Traffic & Rater2 & Machine & 0.537 & 0.426 & 0.445 \\
Traffic & R1\&R2 & Machine & 0.503 & 0.418 & 0.433 \\
Traffic & Rater1 & Machine & 0.470 & 0.411 & 0.419 \\
&&&&&\\
Visibility & Rater2 & Rater1 & 0.620 & 0.624 & 0.579 \\
Visibility & Rater1 & Rater2 & 0.624 & 0.620 & 0.579 \\
Visibility & Rater1 & Machine & 0.603 & 0.606 & 0.577 \\
Visibility & R1\&R2 & Machine & 0.563 & 0.570 & 0.561 \\
Visibility & Rater2 & Machine & 0.523 & 0.547 & 0.531 \\
\hline
\end{tabular}
\end{table}

For the stop-sign presence task, inter-rater agreement was high, with Raters 1 and 2 achieving identical F1-scores of 0.964. When evaluated against either rater or their majority vote, the model yielded F1-scores in the range 0.783–0.803. The precision values for the model was in the range 0.838–0.863 and recall 0.749–0.767.

In the traffic density task, the inter-rater performance was lower, with both directions yielding an F1-score of 0.663. Machine comparisons to individual raters and their majority votes resulted in F1-scores between 0.419 and 0.445, respectively. In all cases, the recall was lower than the precision.

As regards the visibility condition task, Raters 1 and 2 achieved F1-scores of 0.579.
The model demonstrated a comparable performance, with F1-scores in the range 0.531–0.577 based on the reference rater.
The precision and recall values were balanced across all comparisons, thereby indicating a consistent but moderate performance in the human and model evaluations.

\subsection{Model Response Stability}
\begin{figure}[t]
    \centering
    \includegraphics[scale=0.35]{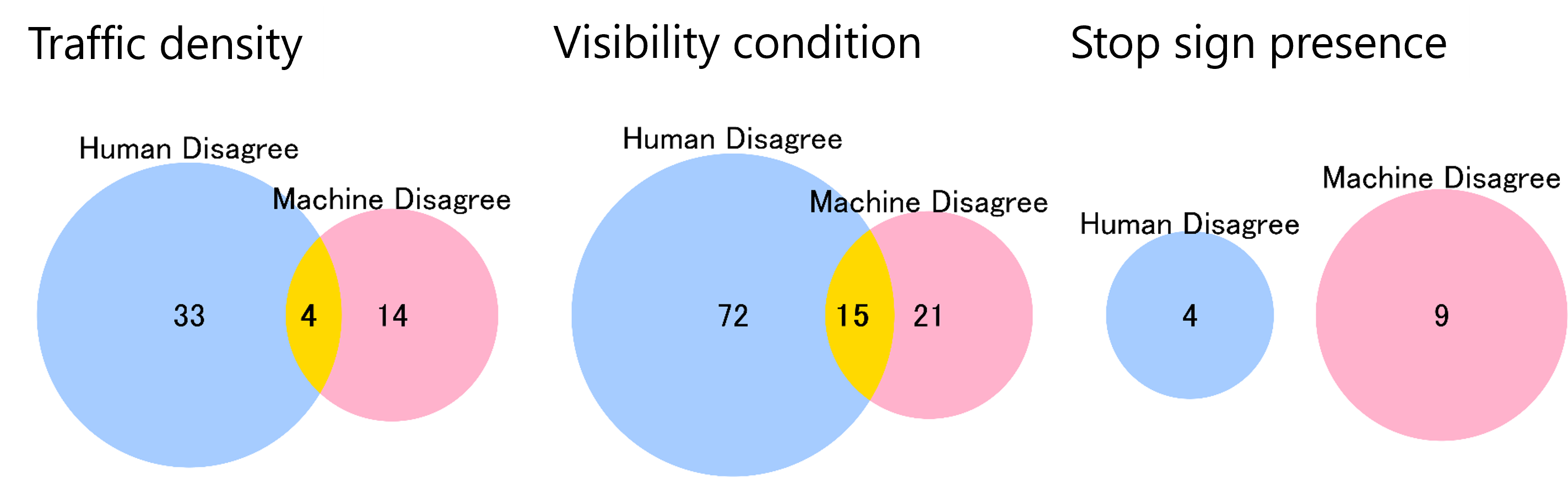}
    \caption{Venn diagrams showing the overlap between human labeling inconsistency and model output variability across three classification tasks. The blue and pink circles represent images with inconsistent labels by human raters or across repeated model outputs, respectively. The yellow region indicates images where both inconsistencies occurred. Circle size denotes the number of images.}
    \label{fig:venn}
\end{figure}

To evaluate the internal consistency, we compared the number of images for which either human raters or the model produced inconsistent classifications.
Figure \ref{fig:venn} illustrates the number of cases for each classification task.

For the traffic density task, the two human raters assigned different labels to 37 of the 200 images, thereby indicating inter-rater inconsistencies. Contrarily, ChatGPT generated inconsistent outputs across ten repeated runs for 18 images. Among these, 4 images overlapped with the human-inconsistent set, thereby suggesting that the model and human raters faced challenges with generating stable classifications for a similar scene.

For the visibility condition task, the human raters exhibited inconsistencies in 87 images, while the model was inconsistent in 36 cases. Fifteen images were common to both sets, thereby indicating a substantial subset of scenes wherein both humans and the model failed to maintain a stable judgments.
This suggests that visibility assessments are sensitive to ambiguities in scene interpretations.

For the stop-sign presence task, inconsistencies between the human raters occurred in only four images, thereby indicating an elevated inter-rater consistency.
Conversely, the model produced inconsistent predictions in nine cases, with no overlap between the human and model inconsistencies. This result implies that while human annotations were highly stable, the model occasionally produced inconsistent outputs, even for scenes wherein humans made consistent labeling.

\section{DISCUSSION}
This study investigated the ability of a multimodal LLM to assess traffic situations using static dashcam images in a manner similar to human judgment. We focused on three factors that are associated with intersection-related risks in elderly drivers: traffic density, intersection visibility, and presence of stop signs. In the stop-sign detection task, the model achieved high precision (above 85\%) but a relatively low recall, thereby indicating a conservative output tendency—--it responded primarily when confident. Similar trends were observed in traffic density and visibility tasks, wherein inter-annotator agreement was low and model predictions showed greater variability. The prompt design considerably influenced the performance of the model: few-shot and multi-shot prompting considerably improved recall in the visibility task, thereby highlighting the significance of contextual guidance. In numerous cases, the labels of the model closely corresponded with their explanation texts, thereby suggesting that decisions are based not only on visual features, but also semantic reasoning. These findings indicate that LLMs can partially reflect human-like scene interpretation, with their performance being highly dependent on the prompt design.

Our findings correspond with those of recent studies on LLMs for diagnostic applications. In the medical field, Ren et al. evaluated ChatGPT-4 for osteosarcoma detection using X-rays \cite{ren2024exploring}. Their model achieved high precision but low recall, and often missed abnormal cases. They attributed this to a cautious response style and limited contextual understanding, thereby cautioning against using the model in clinical settings without safeguards. This corresponds with our findings in the stop-sign task. In the driving domain, Zhang et al. found that combining a visual LLM with pose estimation and CoT prompting improves behavioral labeling \cite{zhang2024integrating}. The recall remained low under zero-shot prompting. However, few-shot prompting with structured examples and stepwise reasoning resulted in considerable improvements. Charoenpitaks et al. showed that structured visual inputs and guided inference improved the natural language descriptions of traffic hazards \cite{charoenpitaks2024exploring}. Their emphasis on structured input and clear prompts corresponded with our findings, particularly in the visibility task.

We also evaluated output consistency by analyzing multiple responses for the same 200 images. While predictions were generally stable, fluctuations were observed, notably in the traffic density and visibility tasks. These variations often coincided with cases wherein humans also disagreed. Such “borderline” scenes included ambiguous visibility owing to lighting or the presence of vehicles that did not block the driving path. These results suggest that both humans and the model faced challenges with ambiguous scenes. Conversely, some errors were consistent across trials, thereby indicating that reproducibility did not guarantee correctness. These findings have several practical implications for future research. In high-precision tasks, the model outputs may identify clear risks. However, in lower-recall tasks, omissions are likely to occur, thereby necessitating human oversight. Notably, fluctuations in the output may highlight scenes that are difficult to assess, and visualizing these cases could enable diagnostic experts in focusing on areas that require closer evaluation. Furthermore, the explanations of the model, often consistent with its labels, enhance transparency and support driver feedback and training. As noted by Alzahem et al. \cite{alzahem2023unlocking}, interpretability remains a key strength of LLMs in applied settings.

\subsection{Limitations}
This study had several limitations. First, the dataset was limited (200 images) and primarily used for prompt engineering. Future studies should test for generalizability using larger and independently sampled datasets. Annotations were conducted using only two raters, which may have introduced individual bias. Broader annotation efforts, including comparisons between expert and non-expert raters, are required to improve reliability. Although our study aimed to support elderly driver assessment, we did not directly examine how the outputs of the model correlate with actual risky behaviors. Future work should compare the model predictions with naturalistic driving data. Finally, we only evaluated ChatGPT-4o. Newer models, such as o1 and o3 offer different architectures and reasoning mechanisms, and their comparison in similar tasks could provide valuable insights.

\section{CONCLUSION}
This preliminary study suggests that multimodal LLMs, when guided by well-designed prompts, can approximate human-like judgments in traffic scene interpretation, particularly for tasks relevant to elderly driver assessments such as visibility, traffic density, and stop-sign recognition. These findings highlight the potential of LLMs as supportive tools for scene-level driving-risk assessments. Future studies should focus on validating this approach by using larger datasets, diverse driving scenarios, and newer model architectures.

\bibliography{mybibfile}

\end{document}